\documentclass[journal]{IEEEtran}
\ifCLASSINFOpdf
\else
\fi

\usepackage{booktabs} 
\usepackage{array}    
\usepackage{bbding}
\usepackage{graphicx}
\usepackage{listings}
\usepackage{xcolor}
\usepackage{url}
\usepackage[colorlinks, linkcolor=blue]{hyperref}

\lstdefinestyle{Python}{
    language        =   Python, 
    basicstyle      =   \zihao{-5}\ttfamily,
    numberstyle     =   \zihao{-5}\ttfamily,
    keywordstyle    =   \color{blue},
    keywordstyle    =   [2] \color{teal},
    stringstyle     =   \color{magenta},
    commentstyle    =   \color{red}\ttfamily,
    breaklines      =   true,  
    columns         =   fixed,  
    basewidth       =   0.5em,
}

\begin{document}
%
\title{VeryFL: A Verify Federated Learning Framework Embedded with Blockchain}
%
%
%

\author{Yihao Li,
        Yanyi Lai,
        Chuan Chen*,~\IEEEmembership{Member, IEEE,}
        and Zibin Zheng,~\IEEEmembership{Fellow, IEEE}}
%
%

\markboth{Journal of \LaTeX\ Class Files,~Vol.~14, No.~8, August~2015}%
{Shell \MakeLowercase{\textit{et al.}}: Bare Demo of IEEEtran.cls for IEEE Journals}
%



\maketitle

\begin{abstract}
Blockchain-empowered federated learning (FL) has provoked extensive research recently. Various blockchain-based federated learning algorithm, architecture and mechanism have been designed to solve issues like single point failure and data falsification brought by centralized FL paradigm. Moreover, it is easier to allocate incentives to nodes with the help of the blockchain. Various centralized federated learning frameworks like FedML, have emerged in the community to help boost the research on FL. However, decentralized blockchain-based federated learning framework is still missing, which cause inconvenience for researcher to reproduce or verify the algorithm performance based on blockchain. Inspired by the above issues, we have designed and developed a blockchain-based federated learning framework by embedding Ethereum network. This report will present the overall structure of this framework, which proposes a code practice paradigm for the combination of FL with blockchain and, at the same time, compatible with normal FL training task. In addition to implement some blockchain federated learning algorithms on smart contract to help execute a FL training, we also propose a model ownership authentication architecture based on blockchain and model watermarking to protect the intellectual property rights of models. These mechanism on blockchain shows an underlying support of blockchain for federated learning to provide a verifiable training, aggregation and incentive distribution procedure and thus we named this framework VeryFL (A \underline{Ver}if\underline{y} \underline{Federated Learninig} Framework Embedded with Blockchain) . The source code is avaliable on \href{https://github.com/GTMLLab/VeryFL}{https://github.com/GTMLLab/VeryFL}
\end{abstract}


%
\IEEEpeerreviewmaketitle

\section{Introduction}
%
%
%
%

\IEEEPARstart{F}{ederated} learning is a distributed machine learning paradigm that allows all participants to collaboratively train machine learning models from multiple data sources without disclosing private data. Federated learning offers a solution to challenges associated with data silos and privacy preservation, presenting a wide spectrum of potential applications. In recent years, federated learning has been a popular research field in machine learning, with numerous significant works focusing on heterogeneity \cite{li2020federated, karimireddy2020scaffold, li2021model}, communication efficiency \cite{rothchild2020fetchsgd, wang2022progfed}, privacy protection \cite{zhao2022pvd, sun2020ldp} and other difficult problems. However, federated learning under centralized server still face the risk of single point failure, data falsification and lack of incentives. Moreover, federated learning model which is trained by multiple parties is exposed to risks such as model illegal copying, misuse and free-riding\cite{li2022fedipr}, and thus need a mechanism to protect intellectual property rights. To face both of the risks brought by centralized paradigm and model rights, blockchain has become a novel solutions for FL to solve these challenges and build a more robust and secure execution environment.

\newcolumntype{C}[1]{>{\centering\arraybackslash}m{#1}} 
\newcolumntype{L}[1]{>{\raggedright\arraybackslash}m{#1}} 

\begin{table*}[htbp]
\label{table:other_tools}
\centering
\setlength{\tabcolsep}{5pt} 
\caption{Comparison with existing federated learning open-source frameworks}
\normalsize 
\begin{tabular}{@{}L{5cm}C{3cm}C{3cm}C{3cm}C{3cm}@{}}
\toprule
& \textbf{FL Benchmark} & \textbf{Blockchain Embedded} & \textbf{FL-side Architecture} & \textbf{Model Authentication} \\ \midrule
\hspace{5pt}EasyFL \cite{zhuang2022easyfl}                & \Checkmark & \textbf{-} & \Checkmark & \textbf{-} \\
\hspace{5pt}FedML \cite{he2020fedml}              & \Checkmark & \textbf{-} & \Checkmark & \textbf{-} \\
\hspace{5pt}GFL \cite{mindspore} & \textbf{-} & \Checkmark & \textbf{-} & \textbf{-} \\
\hspace{5pt}FedCoin \cite{liu2020fedcoin}             & \textbf{-} & \Checkmark & \textbf{-} & \textbf{-} \\
\hspace{5pt}blocklearning \cite{blocklearning}             & \textbf{-} & \Checkmark & \textbf{-} & \textbf{-} \\
\hspace{5pt}FLoBc \cite{ghanem2022flobc}               & \textbf{-} & \Checkmark &\textbf{-} & \textbf{-} \\
\hspace{5pt}\textbf{VeryFL}      & \textbf{\Checkmark} & \textbf{\Checkmark} & \textbf{\Checkmark} & \textbf{\Checkmark} \\ \bottomrule
\end{tabular}
\end{table*}

\par Currently, the open-source framework of federated learning is mainly based on centralized client-server paradigm. For example, FedML is a large-scale federated learning framework that can be deployed to serve as a federated learning infrastructure in production environment. And for experimental propose, there are many frameworks which contains a lot of benchmark to help reproduce the FL experiment such as EasyFL.
\par However, after investigating the existing blockchain-based FL framework on Github, we found that there is still a lack of the blockchain-based FL framework that provides a convenient experimental environment. Existing frameworks are either a demo to design a specific algorithm or lack of the basic benchmark of FL to execute experiments compared with the centralized FL frameworks. For example, FedCoin builds a p2p payment system for FL, but it lacks the basic FL algorithm and dataset to run other experiment on it. Based on the above investigation, we find that it is necessary to build a unified federated learning framework embedded with blockchain to provide the relative experiment with an execution environment.

\par Next, we will introduce our open-source framework VeryFL based on PyTorch and Ethereum. Different from existing federated learning frameworks, VeryFL extends the support for algorithm concerned with blockchain federated learning and have three main features. (1) VeryFL comes with the basic federated learninig algorithm and benchmark dataset to execute FL experiment like other centralized FL framework. (2) VeryFL embedded with the Ethereum network which is interacted through the python SDK. Many on-chain mechanism can be ealisy implemented with smart contract and verified in real blochchain environment. (3) VeryFL designs algorithm to provide model ownership verification services to protect model rights through blockchain. Smart contracts are introduced to assist in model authentication and are responsible for the distribution, management and identification of model tokens. 

We summarize the characteristics of the existing mainstream open-source frameworks alongside the VeryFL framework in Table \ref{table:other_tools}.

The characteristics of VeryFL are delineated as follows:
\begin{itemize}
    \item \textbf{Usability:} VeryFL offers benchmark datasets and aggregation algorithms for federated learning, requiring minimal configuration in the absence of specialized demands, thereby presenting a novice-friendly environment.
    \item \textbf{Scalability:} Owing to its modular design, VeryFL is readily extensible, facilitating users to customize datasets, aggregation algorithms, and other components pertinent to their specific tasks.
    \item \textbf{Blockchain Embedded:} We embedded Ethereum for user to write and invoke on-chain mechanism through smart contract. Beyond providing foundational support for federated learning, VeryFL also affords a prototype system for model ownership verification, thereby broadening the scope of federated learning applications.
    \item \textbf{Model Rights Protection:} We implement our Tokenized Model\cite{li2023token} on VeryFL. With the watermark embedded in the model, we implement a framework where the blockchain turns the model into a non-fungible token(NFT) for the first time to better protect the model rights and allow model to be transacted.
\end{itemize}

In the subsequent sections, we will introduce VeryFL in detail. Section 2 will introduce the architecture and various technologies utilized in VeryFL. Section 3 will introduce the novel attributes of VeryFL. Section 4 will provide practical instances of using VeryFL. Finally, Section 5 will be reserved for summarization and prospective outlooks.

\section{Framework}

\begin{figure}
	\centering
	\includegraphics[width=0.48\textwidth]{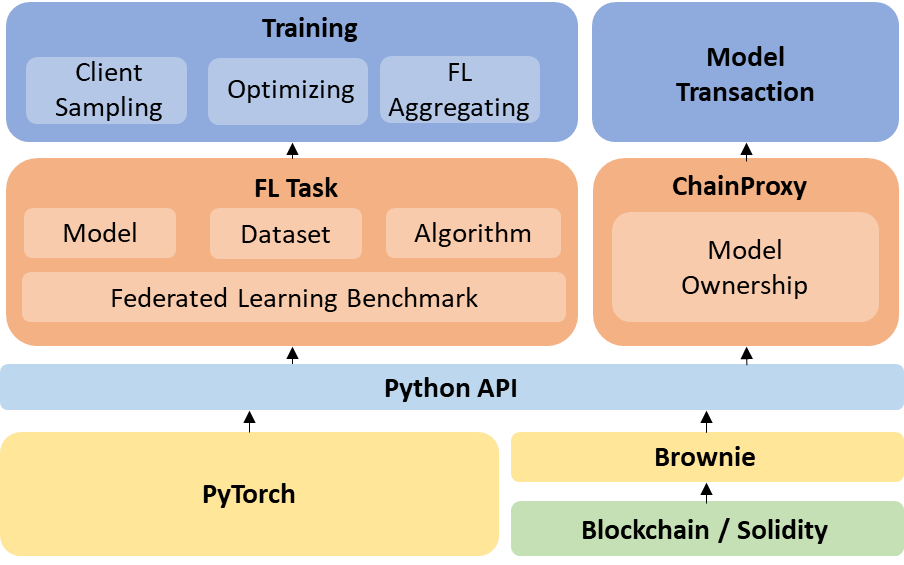}
	\caption{Framework of VeryFL.}
	\label{fig: framework}
\end{figure}

In this section, we will introduce the overall framework of VeryFL constructed upon PyTorch and Ethereum. The framework is shown in Fig. \ref{fig: framework}. With the help of Brownie SDK, the blockchain (smart contracts) provides Python API together with PyTorch at the foundational layer. Atop the Python API, we encapsulate the federated learning process into a \emph{Task} class to perform FL training and we wrap the blockchain API into \emph{ChainProxy} class to deal the interaction with the smart contract. Therefore, the architecture can be roughly divided into two principal modules: the blockchain component and the federated learning task component.

\subsection{Blockchain}

Blockchain often acts as a de-trusted ledger among FL training nodes to provide a unchangeable record of the FL training process. Therefore, the blockchain module is mainly responsible for the management and election of the client, recording of the training process, and distribution of the incentives. In the blockchain module, smart contracts are executed on the Ethereum blockchain network using Solidity, subsequently interacting with PyTorch via the Python SDK, Brownie. Ethereum is an programmable blockchain platform, designed to facilitate smart contracts and decentralized applications. Solidity is a programming language tailored specifically for Ethereum smart contracts, while Brownie is a Python-based development framework for the development and testing of Ethereum smart contracts. These components together form a comprehensive ecosystem.

In VeryFL, blochchain module acts as an manager of the training network. If a user wishes to participate in federated learning training, it is first required to apply for an accounts id for user management. For model copyright protection, blockchain will distribute watermark for each unique model, serving as an identifier for model copyright. This token is bound to the model and is recorded in the blockchain, specifying the user to whom it belongs. And for training management, client and server will upload per-round training result including accuarcy, loss and dataset size to the blockchain for client selection and incentive distribution. Through this bottom blockchain module, many on-chain mechanism can be implemented on the Ethereum network.
\subsection{Federate Learning Task}

\par In common blockchain-FL, all node has the equal opportunity to perform aggregation and no specific defined server. However, to keep the compatibility with the centralized FL, VeryFL still defines the \textit{Aggregator} class to perform the aggregation task. With the election of the blockchain during each trainig round, we handle this aggregator to different client and simulate the decentralized aggregation process. 

\par The main entry of the FL task is in \textit{Task} object. It encapsulates necessary components for a federated learning process and controls the training logic where the FL training process starts. With the passed-in-parameters, \textit{Task} will initialized the component like models, datasets and federated learning algorithms. VeryFL provides some benchmarks dataset and models and some embedded FL algorithm, facilitating a rapid construction of a federated learning workflow. In addition to benchmarks provided by VeryFL by default, users can easily customize these components according to their individual requirements by inheriting from the corresponding abstract classes and rewriting the core code of algorithm.

\section{Features}
\subsection{Federated Learning Framework}
\par Similar with all other federated learning frameworks, the framework comes with several common federated learning benchmarks and image classification datasets in order to make it easy for users to quickly start a federated learning experiment. Common federated learning experiments are often conducted on the CIFAR10, CIFAR100, and FashionMnist datasets, and the framework provides pre-configurations in the Benchmark module to get start and running quickly. In addition to the basic configuration and modules, some utils classes are designed to personalize the experiment. For example, \textit{DatasetSpliter} is provided at the dataset partition to support federated learning experiments with Non-IID data distribution. 
\par 
\subsection{On-chain Mechanism with Smart Contract}
With the help of the programmable Ethereum, VeryFL can implement the on-chain mechanism or algorithm through smart contract. With the help of brownie python API, VeryFL can call these predefined smart contract during training process. Many research on blockchain-FL focus on building reliable, private and secure FL system. This feature provide us with the convenience to test our on-chain FL algorithm in a real blockchain network. VeryFL implement some example smart contract to execute the client registration, client election and training result record functions.

\subsection{Model Ownership verification with Blockchain}
Beyond the FL training functions of VeryFL, we also build a model rights protection and transaction platform on VaryFL as an example of blockchain empowered federated learning. The basic algorithm references the model watermarking techology represented by FedIPR\cite{li2022fedipr}. The proposal of combining FedIPR with blockchain model watermarking algorithm is based on the need for model ownership authentication, where by embedding specific watermarking information, the trainer can claim ownership of the model. Different with the FedIPR scenario, the framework hands over the distribution, recording and authentication process of watermarks to smart contracts to manage, making the distribution process of watermarks more trustworthy. Furthermore, through this process, VeryFL establish a model-user-watermark mapping, where the model trained by the user itself is recorded on the chain as an on-chain asset similar to NFT through the watermark as a representative. In this way, emulating the nature of NFT, we can implement marketable behaviors such as trading and ownership verification against deep learning models. A detailed description of the above process is documented in our previous papers where we propose the tokenized model\cite{li2023token}.

\section{Code Practice}

\subsection{Modulized Design}
In order to facilitate the subsequent expansion of the federated learning algorithm, most parts of the framework have adopted modular design. Decoupling each parts of the federated learning makes it possible for VeryFL to contains different FL algorithm. With the predefined benchmark's configure file as the entry point which is passed into \textit{Task} object, the server, the model, the dataset, the client and the trainer, are instantiated as a single object and assembled together for the training of federated learning. The advantage of implementing each part as a module is that framework can reuse the same parts of the federated learning algorithm as much as possible. For instance, FedAvg and FedProx only differ in client training, the same \textit{Aggregator} module can be reused on both algorithms. While in other cases, implementing a federated learning algorithm requires implementing both \textit{Aggregator} and \textit{Trainer}. By adopting modular design, we can easily extend new FL algorithm into VeryFL.


\subsection{Task Start}
To facilitate a quick start of a task, by importing the predefined training parameters and passing it into Task, we can quickly start a federated learning training task.
\begin{lstlisting}[language=Python]
from task import Task
import config.benchmark
global_args, train_args, algorithm = benchmark.get_args()
classification_task = Task(global_args=global_args, train_args=train_args, algorithm=algorithm)
classification_task.run()
\end{lstlisting}
VeryFL provides benchmark for FL tasks. In \textit{Benchmark} class, VeryFL devided the arguments of the FL training into \textit{global\_args}, \textit{train\_args} and \textit{Algorithm}. The \textit{global\_args} controls the overall setting of a FL task like model, datasets and number of clients. The \textit{train\_args} controls the traing arguments in local model training like learning rate, optimizer and weight decay. By changing the parameters in \textit{Benchmark}, VeryFL can start a customized training.

\subsection{Blockchain Interface}
Since blockchain development is almost inevitably accompanied by the writing of Solidity, VeryFL provides \textit{ChainProxy} as the interface to interact with the blockchain. Users can deploy smart contracts written by themselves and write wrapper functions in \textit{ChainProxy} to wrap the smart contracts interface into python functions that can work together with PyTorch. 
\subsubsection{Start Ethereum Network}

\par The following code line 1 shows the process of starting the blockchain network, which start an instance of a blockchain network. After starting the network successfully, VeryFL begins to deploy the smart contracts.

\subsubsection{Deployment of Smart Contracts}
\par At the begining, the blockchain will have ten accounts whose addresses is stored in a list. We use accounts[0] as the server accounts to deploy smart contracts. The deployment procedure is shown in line 4, 5. The server accounts  After deployment, VeryFL can call functions through the object of a  smart contract instance.
\subsubsection{Predefined Smart Contract}
VeryFL provides some predefined smart contract which have already been wrapped in \textit{ChainProxy}. They provides the registration function that corresponds the client to the blockchain user's address, the training result recording function and the model watermark function mentioned in Chapter 3. These processes can be called to supervise and manage the FL training.

\begin{lstlisting}[language=Python]
from brownie.project.chainServer import *
network.connect('development')
server_accounts = accounts[0]
watermarkNegotiation.deploy({'from':server_accounts})
clientManager.deploy({'from':server_accounts})
\end{lstlisting}

\section{Future Work}
This report present an open-source framework VeryFL. In this framework, we design an extensive and flexible workflow for FL. Besides, we also embed blockchain into this framework to help boost the research of blockchain-empowered federated learning. With the help of embedded Ethereum network, we design a prototype system for model ownership verification mechanism and the incentive mechanism as an example of combining blockchain with FL. We will reproduce more blockchain-based algorithm on this platform to help user verify these algorithm conveniently.


%




\ifCLASSOPTIONcaptionsoff
  \newpage
\fi



%

%

\bibliographystyle{IEEEtran}
\bibliography{IEEEabrv,reference}








\end{document}